\documentclass[letterpaper]{article} %
\usepackage{aaai25}  
\usepackage{times}  
\usepackage{helvet} 
\usepackage{courier} 
\usepackage[hyphens]{url} 
\usepackage{graphicx} 
\usepackage{natbib} 
\usepackage{caption} 
\usepackage{algorithm}
\usepackage{algorithmic}
\usepackage{arydshln} 
\usepackage{enumitem}
\usepackage{xcolor}
\usepackage{amsmath}
\usepackage{threeparttable}
\usepackage{multirow}
\usepackage{placeins}
\usepackage{float}

\usepackage{amsmath}
\usepackage{amssymb}
\usepackage{xcolor} 

\newcommand{\figref}[1]{Figure~\ref{#1}}
\newcommand{\tableref}[1]{Table~\ref{#1}}

\usepackage{newfloat}
\usepackage{listings}
\DeclareCaptionStyle{ruled}{labelfont=normalfont,labelsep=colon,strut=off} 
\floatstyle{ruled}
\newfloat{listing}{tb}{lst}{}
\floatname{listing}{Listing}

\title{Data-driven Precipitation Nowcasting Using Satellite Imagery}
\author{
    Young-Jae Park\textsuperscript{\rm 1},
    Doyi Kim\textsuperscript{\rm 2}, 
    Minseok Seo\textsuperscript{\rm 2}, 
    Hae-Gon Jeon\textsuperscript{\rm 1,3}\thanks{Corresponding authors} and 
    Yeji Choi\textsuperscript{\rm 2}\textsuperscript{$*$}
}
\affiliations{
    \textsuperscript{\rm 1}GIST\quad
    \textsuperscript{\rm 2}SI Analytics\quad 
    \textsuperscript{\rm 3}Yonsei University
}

\usepackage{bibentry}

\begin{document}

\maketitle

\begin{abstract}

Accurate precipitation forecasting is crucial for early warnings of disasters, such as floods and landslides. Traditional forecasts rely on ground-based radar systems, which are space-constrained and have high maintenance costs. Consequently, most developing countries depend on a global numerical model with low resolution, instead of operating their own radar systems. To mitigate this gap, we propose the Neural Precipitation Model (NPM), which uses global-scale geostationary satellite imagery. NPM predicts precipitation for up to six hours, with an update every hour. We take three key channels to discriminate rain clouds as input: infrared radiation (at a wavelength of 10.5 $\mu m$), upper- (6.3 $\mu m$), and lower- (7.3 $\mu m$) level water vapor channels. Additionally, NPM introduces positional encoders to capture seasonal and temporal patterns, accounting for variations in precipitation. Our experimental results demonstrate that NPM can predict rainfall in real-time with a resolution of 2 km. The code and dataset are available at \url{https://github.com/seominseok0429/Data-driven-Precipitation-Nowcasting-Using-Satellite-Imagery}.

\end{abstract}

\section{Introduction}
As global warming accelerates, the damage caused by natural disasters is on the rise. Particularly with increasing temperatures, the intensity of extreme precipitation events escalates~\cite{ombadi2023warming}, leading to significant human casualties due to disasters such as floods, landslides, and soil erosion. To mitigate the loss of life from these disasters, accurate and real-time precipitation forecasting is essential.

Traditionally, precipitation forecasting relies on various observational equipment such as radar systems and numerical weather prediction (NWP) models. For instance, the HRRR model~\cite{dowell2022high} utilizes radar data, satellite data, surface observations, aircraft data, weather buoys and ships, model initialization data, and fire/smoke data to provide forecasts at a resolution of approximately 3 km. Additionally, global NWP models such as ECMWF Reanalysis v5 (ERA5)~\cite{hersbach2020era5} and integrated forecasting system
 (IFS)~\cite{wedi2015modelling} perform forecasts at a resolution of around 25 km.

\begin{figure}[t]
\centering
  \includegraphics[width=1.0\linewidth]{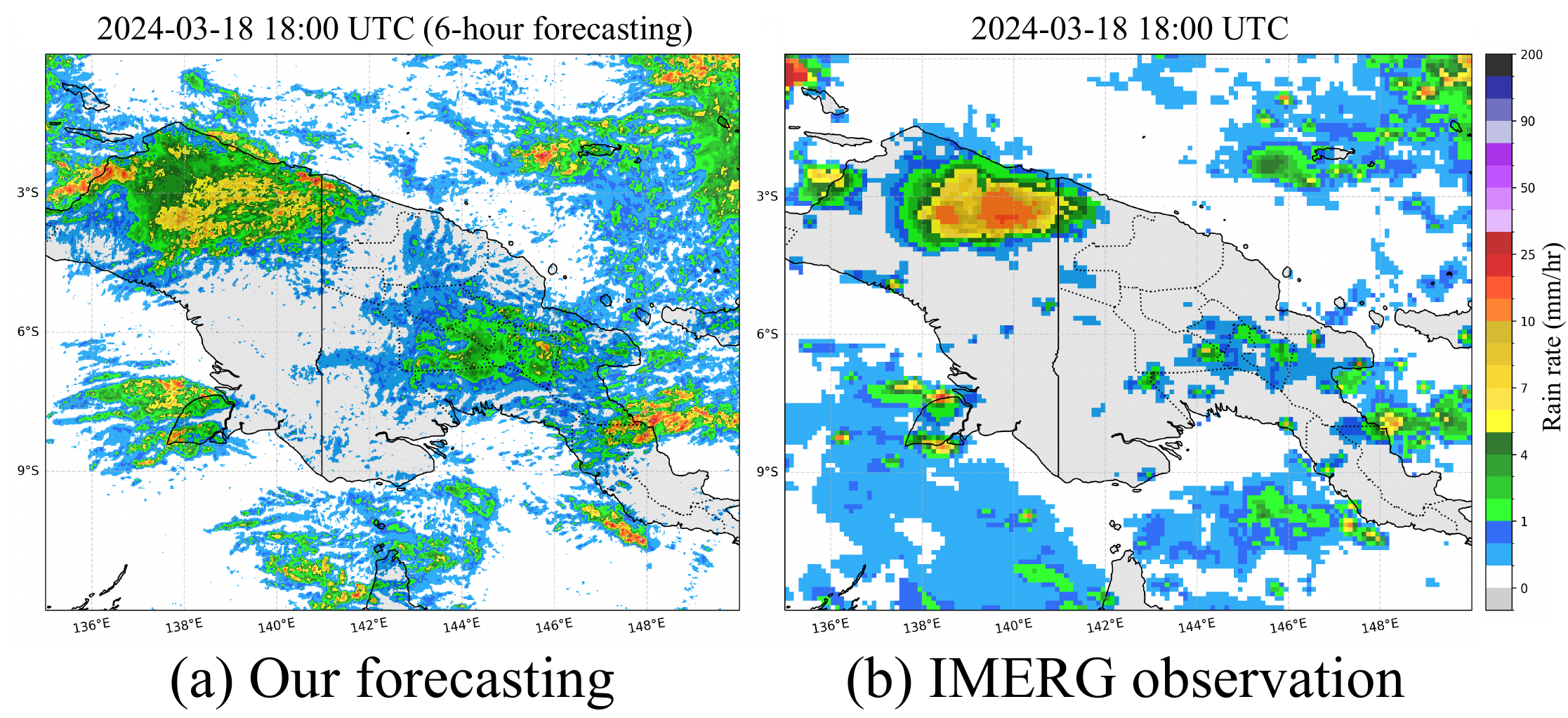}
  \caption{{(a) Our +6 hour forecasting results without radar. (b) NASA IMERG observation for March 18, 2024, Papua New Guinea flood case. (Note that NASA GPM IMERG-late run is accessible only 14 hours after observation.)}
 }
\label{fig:NPM_tea}
\end{figure} 

Despite advancements in the observational equipment and the NWP models, essential tools for precipitation forecasting involve installation and maintenance costs that can reach about billions of dollars. Furthermore, algorithms like IFS and ERA5 require supercomputers, making them challenging to operate in countries with limited budgets and resources.

To overcome these issues, numerous data-driven weather forecasting methods have been proposed. Pangu-Weather~\cite{bi2023accurate} and GraphCast~\cite{lam2023learning} demonstrate superior performance compared to traditional NWP models like IFS, even when running in a single GPU environment. Similarly, a series of MetNet~\cite{andrychowicz2023deep, sonderby2020metnet} shows high accuracy in precipitation forecasting while also operating efficiently on a single GPU. However, these global weather forecasting models still rely on NWP data for their initial conditions, meaning they have not fully eliminated the dependency on supercomputers. Additionally, with a resolution of 25 km, they remain insufficient for predicting localized heavy rainfall events, such as flash floods.

The MetNet series and various radar-to-radar benchmark~\cite{veillette2020sevir} models take radar data as input, and produce radar-based outputs, making radar infrastructure essential. Moreover, radar-to-radar models are unable to detect developing precipitation types that do not appear in radar signals yet, further reducing their effectiveness in certain situations.

\begin{figure*}[!t]
\centering
  \includegraphics[width=1.0\linewidth]{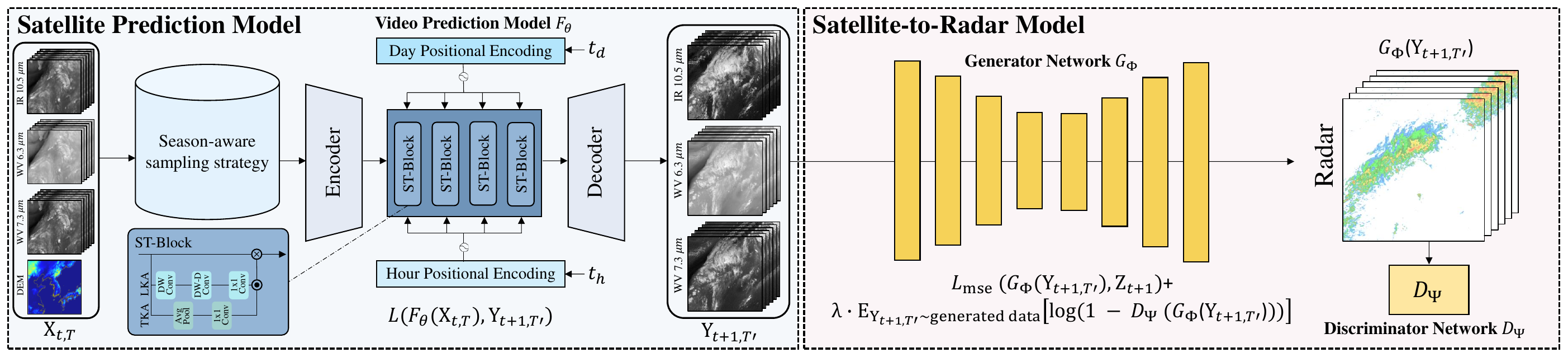}
  \caption{{Overview of NPM Architecture. First, the Satellite Prediction Model takes season-aware sampled satellite sequences and then predicts future frames. Second, the Satellite-to-Radar Model generates precipitation from predicted satellite sequences.}
 }
\label{fig:overview_1}
\end{figure*}

To address the challenges of supercomputer dependency, radar-only systems, and the difficulty of predicting precipitation without prior radar signals, we propose the real-time neural precipitation model (NPM). In addition, to train our NPM model, we introduce the Geostationary Satellite-to-Radar dataset, which is called Sat2Rdr.
Our approach is inspired by traditional methods of predicting precipitation using atmospheric states and cloud characteristics from satellite imagery, which have historically relied on the correlation between cloud-top brightness temperature and surface rainfall rates \cite{arkin1979relationship, sorooshian2000evaluation, huffman2010trmm}.
The dataset utilizes the Infrared (IR) channel at 10.5 $\mu m$, the Water Vapor (WV) channel at 6.3 $\mu m$, and the WV channel at 7.3 $\mu m$—channels closely linked to precipitation in traditional algorithms.

Our NPM consists of two stages:
The first stage predicts satellite images showing the formation and dissipation of clouds associated with precipitation. 
The second stage estimates rainfall from the predicted satellite images, accurately predicting rain rates by analyzing cloud type and growth stage through infrared and water vapor channels.
Since the NPM relies solely on satellite imagery, it does not inherently predict seasonal or diurnal precipitation patterns.
To overcome this, we incorporate day and hour positional embeddings into the NPM, allowing it to learn these patterns.
As shown in ~\figref{fig:NPM_tea} from the Papua New Guinea flood case, NPM is capable of forecasting precipitation even in regions without radar coverage, as it uses only satellite imagery and Digital Elevation Model (DEM) data as input.

We validate the effectiveness of our NPM using various video prediction models on the geostationary Sat2Rdr dataset and achieve the best performance over relevant approaches. Additionally, we show an interesting case study that our NPM works well in a North Korean flood event without any radar coverage.

\section{Related Works}
\subsection{Global Weather Forecasting}
Global weather forecasting traditionally relies on NWP models which simulate atmospheric conditions using complex physical equations.
While effective, NWP models suffer from limitations, including high computational costs and a heavy dependency on the precise assimilation of observational data.
Recent advancements in data-driven manners, such as Pangu-Weather~\cite{bi2023accurate}, GraphCast~\cite{lam2023learning}, and LT3P~\cite{park2024long} demonstrate competitive results compared to traditional NWP-based methods, even in single GPU environments without any need for supercomputers.
However, the data-driven global weather prediction models still face significant challenges in forecasting precipitation.
First, these models often depend on NWP data as input, which means that they still require supercomputers for generating the data.
Second, since these models use reanalysis data from NWP models (e.g., ERA5) as ground truth, any biases~\cite{lavers2022evaluation} in the NWP data are likely to be reflected in their outputs, particularly in precipitation forecasts.
Lastly, the spatial resolution of NWP models significantly influences the grid size, where at a resolution of 0.25 degrees (approximately 25 km).
These limitations have been considered as challenges that should be addressed in the field of data-driven weather forecasting.

%

\subsection{Regional Precipitation Forecasting}

Unlike global weather forecasting, regional precipitation forecasting targets predictions with high spatial resolution (e.g., 2 km).
The SEVIR dataset~\cite{veillette2020sevir}, a widely used for precipitation nowcasting, contains 20,393 weather events, each comprising a sequence of radar frames spanning 4 hours with a spatial resolution of 1 km$\times$1 km. The MeteoNet dataset~\cite{meteo2020} covers a vast area of 550 km$\times$550 km in France and includes over three years of observational data. Additionally, the Shanghai Radar dataset~\cite{chen2020deep} is generated through volume scans conducted from 2015 to 2018 in Pudong, Shanghai, covering a spatial area of 501 km$\times$501 km. However, these precipitation benchmark datasets rely heavily on radar data.

To advance the field, several models have been developed. DGMR~\cite{ravuri2021skilful}, a Generative Adversarial Network (GAN)-based model, uses radar observations as input to predict weather over an area of 1,536 km$\times$1,280 km with lead times ranging from 5 to 90 minutes. Prediff~\cite{gao2024prediff}, a state-of-the-art model on the SEVIR dataset, integrates knowledge alignment using conservation laws into a diffusion-based video generation model. Other significant contributions include DGDM~\cite{yoon2023deterministic}, which combines deterministic and stochastic models, and DiffCast~\cite{yu2024diffcast}, which employs residual diffusion—both of which have driven substantial progress in data-driven regional precipitation forecasting.

A significant limitation of these datasets and models is their reliance on radar data, making them inapplicable in regions without radar coverage and ineffective when radar signals are absent.

\section{Method}

\subsection{Satellite-based Precipitation Forecasting}
Given geostationary satellite imagery channels, IR 10.5 $\mu$m, WV 6.3 $\mu$m, WV 7.3 $\mu$m, and DEM data, the satellite-based precipitation forecasting framework aims to predict precipitation levels, 6 hours into the future.
However, directly predicting precipitation from satellite imagery channels is challenging due to the difference in modality between the input (satellite images) and the output (precipitation rates), which makes the use of auto-regressive inference strategies infeasible.

To address this challenge, we propose a two-stage model. The first stage focuses on video prediction, and we define the spatiotemporal predictive learning problem as follows:
Given a sequence of frames $\mathbf{X}_{t,T} = \{\mathbf{x}_i\}_{t-T+1}^{t}$ at time $t$ over the past $T$ frames, the goal is to forecast the next $T'$ frames $\mathbf{Y}_{t+1,T'} = \{\mathbf{y}_i\}_{t+1}^{t+1+T'}$ starting from time $t+1$.
Here, $\mathbf{x}_i$ and $\mathbf{y}_i$ represent individual frames where $\mathbf{x}_i, \mathbf{y}_i \in \mathbb{R}^{C \times H \times W}$, with $C$ as the number of channels, $H$ as the height, and $W$ as the width.
The model, parameterized by $\Theta$, learns a forecasting function $F_{\Theta} : \mathbf{X}_{t,T} \rightarrow \mathbf{Y}_{t+1,T'}$ by leveraging both spatial and temporal correlations. In this context, $F_{\Theta}$ is a model trained to minimize the discrepancy between the predicted future frames and the actual future frames.

According to~\cite{veillette2020sevir}, applying GAN training techniques in satellite-to-radar translation tasks has been shown to improve performance.
Following this approach, the second stage of the model is an image-to-image (I2I) translation task with a GAN combined with an MSE loss.
Given the output $\mathbf{Y}_{t+1,T'}$ from $F_{\Theta}$ in the first stage, the goal is to transform the satellite imagery into radar-based precipitation maps:
\begin{equation}
\hat{\mathbf{Z}}_{t+1} = G_{\Phi}(\mathbf{Y}_{t+1,T'})
\end{equation}
$\mathbf{Z}_{t+1} \in \mathbb{R}^{C' \times H \times W}$ represents ground truth radar precipitation map at time $t+1$.

Here, $G_{\Phi}$ is the generator network in the GAN, with parameters $\Phi$, which learns the mapping from satellite images to radar-based precipitation data.
The generator $G_{\Phi}$ is optimized by combining an adversarial loss with a Mean Squared Error (MSE) loss, ensuring that the generated radar maps are realistic and accurately reflect the input satellite imagery:

\begin{equation}
\begin{split}
\Phi^* = \arg\min_{\Phi} L_{\text{MSE}}(G_{\Phi}(\mathbf{Y}_{t+1,T'}), \mathbf{GT}_{t+1}) + \\
\! \lambda \cdot \mathbb{E}_{\mathbf{Y}_{t+1,T'} \sim \text{generated data}}[\log(1 - D_{\Psi}(G_{\Phi}(\mathbf{Y}_{t+1,T'})))]
\end{split}
\end{equation}
Here, $\lambda$ is a hyperparameter that balances between the adversarial loss and the MSE loss.

\subsection{Season-aware Sampling Strategy}
Our Sat2Rdr dataset consists of 41,637 sequential data points, spanning from September 2019 to July 2024 at 1-hour intervals.
Typically, video prediction models are trained by randomly sampling an index and using a sequence from $\text{index}-t$ as input and $\text{index}+t$ as output.
However, this random sampling approach may result in certain sequences being selected more frequently, leading to a bias in the training data towards specific months or seasons.
An alternative way is to train the model on all possible combinations (e.g., $41,637 - 11$ combinations) in each epoch, but this is computationally inefficient.
Therefore, we propose a simple yet effective Season-aware Sampling Strategy.
First, we partition the input data by year and month, ensuring that each month is represented uniformly. Then, within each selected month, we randomly sample indices for training.
This approach ensures that all months are equally represented during training. Additionally, if the model underperforms for specific months, oversampling can be applied to those months.

Let $\mathcal{D}$ denote the dataset, where $|\mathcal{D}| = 41,637$. We partition $\mathcal{D}$ into subsets $\mathcal{D}_{\text{year}, \text{month}}$ for each year and month. During training, we select samples uniformly from each subset:

\begin{equation}
P(\text{month}) = \frac{1}{12}, \quad \forall \quad \text{month} \in \{1, 2, \dots, 12\}.    
\end{equation}

Within each month, we sample indices $\text{index}$ from $\mathcal{D}_{\text{year}, \text{month}}$ uniformly at random:

\begin{equation}
\text{index} \sim \mathcal{U}(\mathcal{D}_{\text{year}, \text{month}})
\end{equation}
\subsection{Satellite Prediction Model}
\noindent\textbf{Day \& Hour Positional Encoding}
Precipitation can have seasonal and daily patterns. Previous studies reveal that the cloud-top brightness temperature values can show different patterns depending on the season \cite{van1993seasonal, wang2004climatology, song2023long}.
For example, clouds in summer tend to have the lowest brightness temperatures in the Korean Peninsula~\cite{song2023long}, which can cause heavier precipitation than in other seasons.
While NWP model data (e.g., temperature and wind fields) could be used to reflect these seasonal and diurnal precipitation patterns, our model does not utilize NWP data, making it challenging to account for these variations.
To address this challenge, we enable the model to infer the season and time by embedding the day and hour of the last date in the input data as a condition. The Day and Hour embeddings range from 0 to 365 and from 0 to 24, respectively:

\begin{equation}
\resizebox{0.5\textwidth}{!}{
$
PE(\text{x}, k) =
\begin{cases} 
\sin\left(\frac{\text{x}}{10000^{\frac{2i}{d}}}\right), & \text{if } k = 2i, \\
\cos\left(\frac{\text{x}}{10000^{\frac{2i}{d}}}\right), & \text{if } k = 2i+1,
\end{cases}
\quad \text{for } \text{x} = \text{day, hour}
$
}
\end{equation}

where $PE$ denotes the positional encoding for a given day or hour, $i$ is the dimension index, and $d$ is the embedding dimension. After computing the embeddings, they are concatenated and passed through two fully connected layers with GELU activation functions:

\begin{equation}
embed = \text{Linear}(\text{GELU}(\text{Linear}([\text{PE(day)}, \text{PE(hour)}]))).    
\end{equation}

This embedding is then provided as a condition to the model, allowing it to consider seasonal and diurnal variations in precipitation patterns, even without direct NWP data input.

\noindent\textbf{Spatio-temporal Modeling}
To achieve computationally efficient video modeling, as proposed in~\cite{gao2022simvp}, our satellite prediction model also adopts an encoder, translator, and decoder structure. However, as highlighted in~\cite{lam2023learning} and~\cite{gruca2023weather4cast}, large context is crucial in weather and satellite image prediction tasks. To incorporate large context while maintaining an efficient architecture, we integrate the large-kernel attention block from~\cite{guo2023visual}.

Since~\cite{guo2023visual} does not include a temporal axis, we extend the large-kernel attention block to the temporal dimension, proposing a spatio-temporal large-kernel attention block. As shown in~\figref{fig:overview_1}, the spatio-temporal block (ST-Block), consisting of temporal attention and spatial attention, is computed in a decomposed manner.

\noindent\textbf{Temporal Consistency Constraint}
In satellite image forecasting, the continuity between frames is a critical aspect that reflects the natural laws governing physical phenomena.
To model this continuity, we first compute the difference between consecutive predicted frames $\hat{\mathbf{Y}}$ and the actual frames $\mathbf{Y}$:

\begin{equation}
\delta \hat{\mathbf{Y}}_i = \hat{\mathbf{Y}}_{i+1} - \hat{\mathbf{Y}}_i, \hspace{2mm} \delta \mathbf{Y}_i = \mathbf{Y}_{i+1} - \mathbf{Y}_i.
\end{equation}

These differences capture essential information about the continuity of natural phenomena.
To penalize larger changes between these differences, we add the following regularization term, based on the Kullback-Leibler divergence between the forecast and future distributions:

\begin{equation}
L_{\text{reg}}(\hat{\mathbf{Y}}, \mathbf{Y})= \sum_{i=1}^{T'-1} \sigma(\Delta \hat{\mathbf{Y}}_i) \log \frac{\sigma(\Delta \hat{\mathbf{Y}}_i)}{\sigma(\Delta \mathbf{Y}_i)}.
\end{equation}

The final loss function of our model is a linear combination of the mean squared error (MSE) loss and this regularization term:

\begin{equation}
\mathcal{L} = \sum_{i=1}^{T'} \|\hat{\mathbf{Y}}_i - \mathbf{Y}_i\|^2 + \alpha L_{\text{reg}}(\hat{\mathbf{Y}}, \mathbf{Y}),
\end{equation}
where $\alpha$ is a weight that balances the two loss terms. This approach ensures that the model maintains continuity between frames, leading to more accurate predictions.

\subsection{Satellite-to-Radar Model}
The Satellite-to-Radar Model is based on generative models as in ~\cite{ravuri2021skilful} and ~\cite{veillette2020sevir}.
However, as noted in~\cite{ravuri2021skilful} and~\cite{veillette2020sevir}, satellite-to-radar translation cannot be considered as a perfect paired image-to-image translation setting.
Even with the same type of cloud, the results can be inconsistent due to small signals that cannot be detected by satellites.
Additionally, the presence of radar echoes further complicates the situation, making it difficult to achieve a perfect paired setting.
Therefore, we treat this as an unpaired setting and choose the baseline from \cite{wu2024stegogan} accordingly.
It is important to note that various image-to-image translation baselines are available, such as Pix2Pix\cite{isola2017image} for paired settings or BBDM~\cite{li2023bbdm}, which is based on diffusion. In this work, we use the approach from ~\cite{wu2024stegogan}, which has empirically demonstrated the best performance.

\begin{table*}[t!]
\centering
\resizebox{1.7\columnwidth}{!}{%
\begin{tabular}{c|cccccc|cccccc|cccccc}
\hline \hline
          & \multicolumn{6}{c|}{\textbf{CSI 1 mm}} & \multicolumn{6}{c|}{\textbf{CSI 4 mm}} & \multicolumn{6}{c}{\textbf{CSI 8 mm}} \\ \hline
\textbf{Method}    & 1h   & 2h  & 3h  & 4h  & 5h  & 6h  & 1h   & 2h  & 3h  & 4h  & 5h  & 6h  & 1h  & 2h  & 3h  & 4h  & 5h  & 6h  \\ \hline
PhyDNet~\cite{guen2020disentangling}   & 0.37  & 0.32  & 0.28  & 0.25  & \textbf{0.22}  & \textbf{0.18}  & 0.23  & 0.19  & 0.16  & \textbf{0.13}  & 0.09  & 0.07  & 0.08  & 0.05  & 0.03  & 0.01  & 0.00  & 0.00  \\
PredRNNV2~\cite{wang2022predrnn}  & 0.38  & 0.31  & 0.27  & 0.23  & 0.20  & 0.16  & 0.22  & 0.18  & 0.15  & 0.11  & 0.08  & 0.06  & 0.06  & 0.04  & 0.01  & 0.00  & 0.00  & 0.00  \\
SimVP~\cite{gao2022simvp}      & 0.36  & 0.30  & 0.26  & 0.22  & 0.19  & 0.14  & 0.21  & 0.17  & 0.14  & 0.10  & 0.07  & 0.04  & 0.05  & 0.02  & 0.00  & 0.00  & 0.00  & 0.00  \\
SimVP-V2~\cite{tan2022simvpv2}   & 0.35  & 0.29  & 0.24  & 0.20  & 0.17  & 0.13  & 0.20  & 0.15  & 0.12  & 0.09  & 0.05  & 0.03  & 0.04  & 0.01  & 0.00  & 0.00  & 0.00  & 0.00  \\
TAU~\cite{tan2023temporal}        & 0.36  & 0.30  & 0.25  & 0.22  & 0.18  & 0.14  & 0.22  & 0.18  & 0.14  & 0.11  & 0.08  & 0.05  & 0.06  & 0.03  & 0.01  & 0.00  & 0.00  & 0.00  \\
SwinLSTM~\cite{tang2023swinlstm}   & 0.37  & 0.31  & 0.27  & 0.23  & 0.19  & 0.15  & 0.23  & 0.19  & 0.15  & 0.12  & 0.08  & 0.06  & 0.07  & 0.04  & 0.02  & 0.00  & 0.00  & 0.00  \\ \hline
Ours   & \textbf{0.49} & \textbf{0.40} & \textbf{0.35} & \textbf{0.32} & 0.20 & 0.17   & \textbf{0.25} & \textbf{0.21} & \textbf{0.17} & 0.12 & \textbf{0.09} & \textbf{0.08}   & \textbf{0.09} & \textbf{0.07} & \textbf{0.05} & \textbf{0.04} & \textbf{0.03} & \textbf{0.02} \\ \hline \hline
\end{tabular}%
}
\caption{Comparison of CSI performance between Video Frame Prediction models and our model. Note that in our satellite-to-radar framework, video prediction is performed in the first step, followed by image-to-image translation in the second step. This allows for the use of auto-regressive models.}
\label{tab:ex1}
\end{table*}
\section{Experiments}
In this section, we provide a detailed explanation of the Sat2Rdr dataset, discuss the model implementation details and evaluation metrics, and conduct an analysis of the experimental results for lead times and seasonal precipitation.
\subsection{Sat2Rdr Dataset}
\begin{table}[H]
\centering
\tiny
\begin{tabular}{c c c c}
\hline\hline
Sensor & Description & Resolution & Spatial Coverage   \\ \hline
Radar    & Hybrid surface rainfall & 500 m &  1,000 km $\times$ 1,000 km  \\ 
IR 10.5 $\mu m$    & Infrared radiation   & 2 km &1,000 km $\times$ 1,000 km  \\ 
WV 6.3 $\mu m$    & Water vapor channels (upper-level)   & 2 km  & 1,000 km $\times$ 1,000 km  \\ 
WV 7.3 $\mu m$     & Water vapor channels (lower-level) & 2 km & 1,000 km $\times$ 1,000 km  \\  \hline
DEM & Digital elevation model & 2 km& 1,000 km $\times$ 1,000 km  \\
\hline\hline
\end{tabular}
\caption{Description of sensor types in our Sat2Rdr dataset}
\label{tab:datasetinfo}
\end{table}
The Sat2Rdr dataset is constructed at 1-hour intervals from September 2019 to June 2024 using the IR 10.5 $\mu m$, WV 6.3 $\mu m$, and WV 7.3 $\mu m$ channels.
The Sat2Rdr dataset is sourced from the GK2A geostationary satellite. The dataset’s geolocation is mapped using the WGS 84 datum and the Polar Stereographic projection, centered on a true scale latitude of 37.45°N and a central meridian of 126.83°E.
Detailed information about the dataset can be found in~\tableref{tab:datasetinfo}.
To fairly evaluate the model across multiple years and months, the Sat2Rdr dataset uses data from September 2019 to June 2023 for training, with the test dataset spanning from July 2023 to June 2024. The radar includes 10 ground-based observation points that are merged, registered, and matched with the satellite data.

Additionally, to incorporate geographical information, a DEM that provides detailed elevation data is included and spatially aligned with the geostationary satellite dataset.

\subsection{Evaluation Metric} To evaluate the performance of our precipitation prediction model, we utilize the CSI 1 mm, CSI 4 mm, and CSI 8 mm evaluation metrics, as done in~\cite{andrychowicz2023deep}. The CSI (Critical Success Index) is calculated as follows:

\begin{equation}
\text{CSI} = \frac{\text{TP}}{\text{TP} + \text{FP} + \text{FN}},
\end{equation}
where \(\text{TP}\) (True Positives) is the number of correctly predicted precipitation events, \(\text{FP}\) (False Positives) means the number of incorrectly predicted precipitation events (predicted precipitation where there was none), and \(\text{FN}\) (False Negatives) represents the number of missed precipitation events (actual precipitation not predicted).
The CSI value ranges from 0 to 1, where 1 indicates perfect prediction accuracy and 0 means no skill.
For additional POD (Probability of Detection) and FAR (False Alarm Ratio) scores, please refer to the supplementary material.

\begin{figure}[t!]
\centering
  \includegraphics[width=1.0\linewidth]{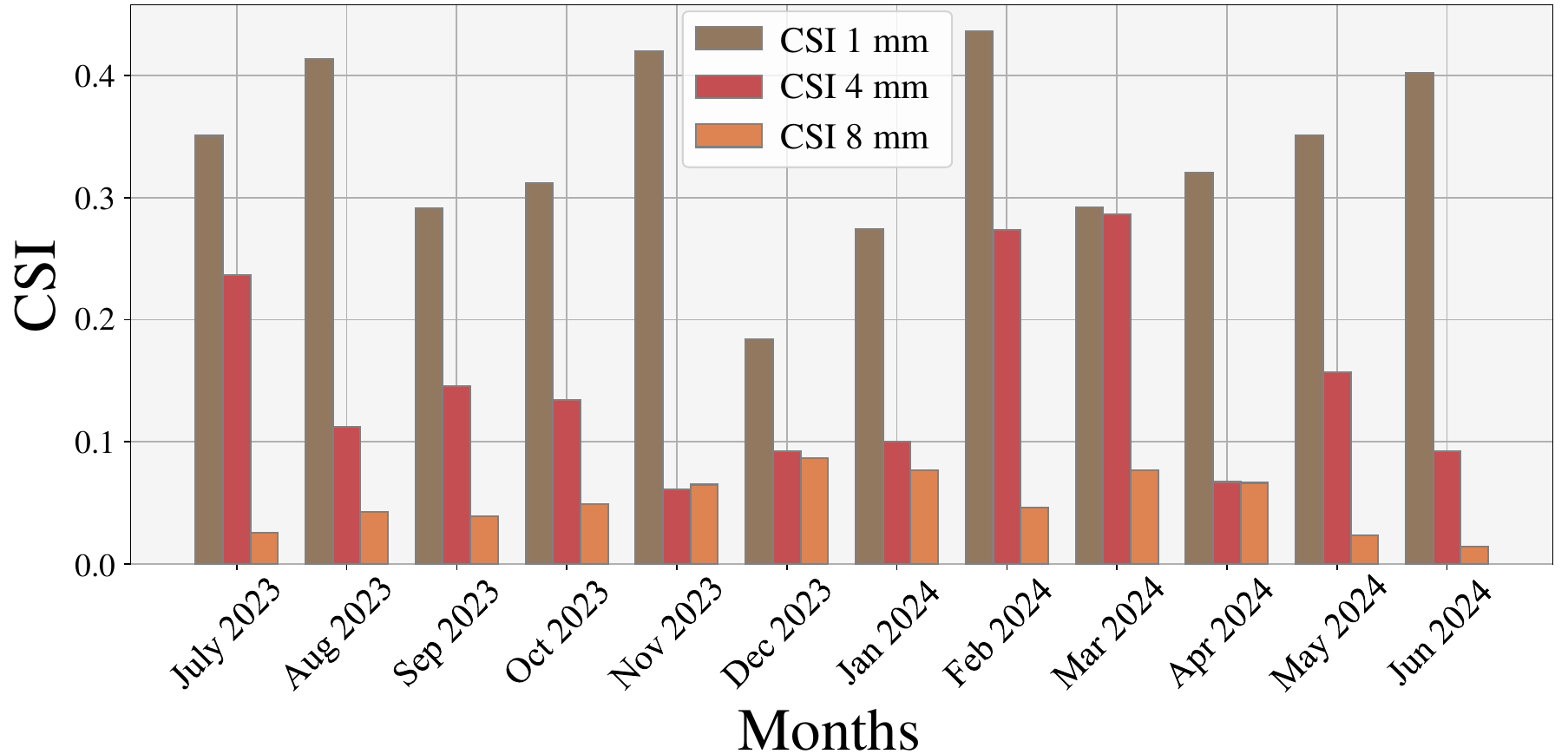}
  \caption{{Performance comparison of CSI 1 mm, CSI 4 mm, and CSI 8 mm by month. Categorical CSI (higher is better). CSI plots are for light (1 mm/h), moderate (4 mm/h), and heavy (8 mm/h) precipitation.}
 }
\label{fig:ex_1}
\end{figure}
\subsection{Baseline}
To fairly evaluate various models on our proposed dataset, we utilize the OpenSTL~\cite{tan2023openstl} video prediction framework.
The OpenSTL framework\footnote{https://github.com/chengtan9907/OpenSTL} provides a fair implementation of both auto-regressive and non-autoregressive models.
In our experiments, we set PhyDNet~\cite{guen2020disentangling}, PredRNNV2~\cite{wang2022predrnn}, SimVP~\cite{gao2022simvp}, SimVP-V2~\cite{tan2022simvpv2}, TAU~\cite{tan2023temporal}, and SwinLSTM~\cite{tang2023swinlstm} as our comparison methods.
Note that all our implementations are performed within the OpenSTL video prediction framework and are planned to be integrated into the official OpenSTL framework repository.
\subsection{Implementation Details}
We train our Sat2Rdr dataset, which has a spatial resolution of 900 $\times$ 900, by randomly cropping it to a size of 768 $\times$ 768. The input timestamp ${t}$ is set to 6, and the output timestamp $\hat{t}$ is also set to 6. The number of encoders and decoders is set to 4, and the number of ST-Blocks is set to 3. The number of channels in the encoders and decoders is set to 64, while the number of channels in the ST-Blocks is set to 512. The weight for the Temporal Consistency Constraint is set to 0.1, and we perform distributed training on 8 NVIDIA A6000 GPUs with a batch size of 1 per GPU. The initial learning rate is set to 1e-4, and cosine learning rate decay is used. Note that during the test phase, we input images of size 900 $\times$ 900 to avoid grid artifacts caused by patch inference. Additionally, all hyperparameters of StegoGAN~\footnote{https://github.com/DAI-Lab/SteganoGAN}~\cite{wu2024stegogan} for our image-to-image translation as a baseline model are set to the default settings, and training is conducted in a paired dataset environment.

\begin{table}[t!]
\centering
\resizebox{0.9\columnwidth}{!}{%
\begin{tabular}{c|c|ccc}
\hline \hline
\textbf{Method}   & \textbf{Sampling Steps}        & \textbf{CSI 1 mm} & \textbf{CSI 4 mm} & \textbf{CSI 8 mm} \\ \hline
Pix2Pix          &  1 &  0.55     &  0.36       &    0.38     \\
StegoGAN         &  1 &  0.54     &  0.31       &    0.29    \\
BBDM             &  200 &  0.66     &  \textbf{0.49}       &    0.43     \\ \hline
StegoGAN (Paired) &  1 &  \textbf{0.66}     &  0.41       &    \textbf{0.51}     \\ \hline \hline
\end{tabular}%
}
\caption{Performance comparison according to image-to-image translation models
}
\label{tab:I2I}
\end{table}

\begin{table}[ht]
\centering
\resizebox{0.8\columnwidth}{!}{%
\begin{tabular}{l|ccc}
\hline \hline
\multicolumn{1}{c|}{\textbf{Component}}   & \textbf{CSI 1 mm} & \textbf{CSI 2 mm} & \textbf{CSI 4 mm} \\ \hline
SimVP                               &   0.245    &   0.121      &   0.011      \\
+LKA                                &   0.260    &   0.121      &   0.010           \\
+TKA                                &   0.264    &   0.125      &   0.020           \\
+Temporal Consistency Constraint    &   0.265    &   0.130      &   0.027           \\
+Day Embedding                    &   0.310    &   0.153      &   0.043           \\
+Hour Embedding                     &   0.316    &   0.151      &   0.049           \\
+Sampling strategy                  &   \textbf{0.321}    &      \textbf{0.153}        &    \textbf{0.050}          \\ \hline \hline
\end{tabular}%
}
\caption{Ablation study on NPM components}
\label{tab:table_aa}
\end{table}

\begin{figure*}[t!]
\centering
  \includegraphics[width=0.85\linewidth]{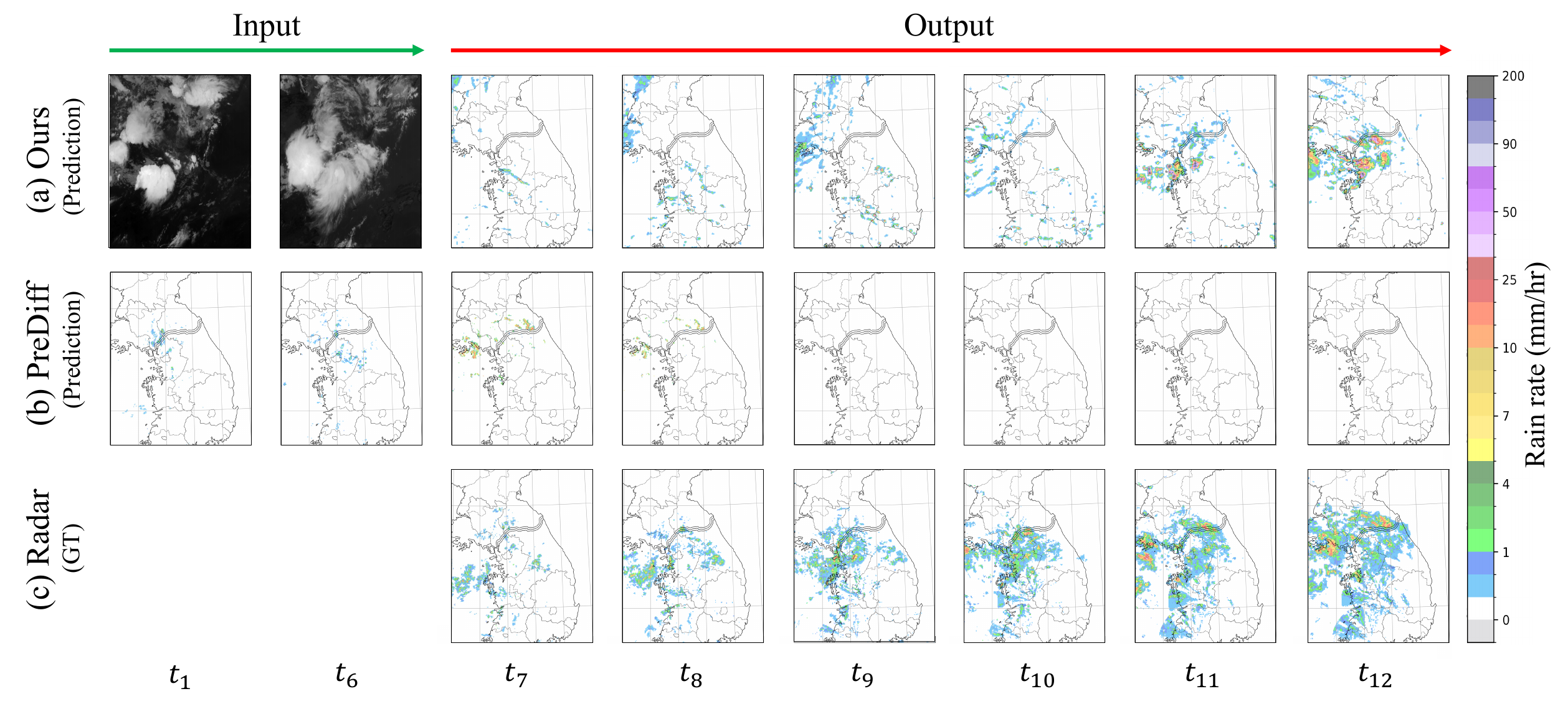}
  \caption{Comparison between the Radar2Radar state-of-the-art method, PreDiff, and our Sat2Rdr approach. (a) shows ours, (b) shows PreDiff, and (c) shows radar-observed ground truths.}
\label{fig:fig_pray}
\end{figure*}
\subsection{Quantitative Results}
\noindent\textbf{Main Results} ~\tableref{tab:ex1} presents a quantitative comparison of various video frame prediction models, including our proposed NPM, using the OpenSTL framework.
As observed, the performance of all models declines as we move from CSI 1 mm to CSI 8 mm, with further decreases as the lead time increases. 
Additionally, NPM consistently outperforms all the baselines in OpenSTL, achieving higher average scores across CSI 1 mm, 4 mm, and 8 mm.

Notably, in the CSI 8 mm scenario, the most models register a CSI score of zero.
Despite these challenges, our model achieves an average CSI score of 0.05 in the 8 mm scenario. This success can be attributed to the inclusion of day encoding in our model, which effectively captures seasonal precipitation patterns.
Unlike moderate rainfall, heavy rain events are highly dependent on seasonal factors, and our approach is more effective at accounting for these conditions.

It is worth noting that our experimental results show a generally lower CSI score than those in~\cite{andrychowicz2023deep}. This discrepancy arises because the study in ~\cite{andrychowicz2023deep} utilizes a combination of radar, NWP, satellite, and DEM data, simultaneously.
Additionally, we do not include radar-to-radar short-term precipitation forecasting models such as ~\cite{gao2024prediff}, ~\cite{yu2024diffcast}, and ~\cite{yoon2023deterministic} in our experiments, as these models are specifically designed for short-term precipitation forecasting. Expanding these models to sat-to-radar nowcasting would require significant modifications. 

\noindent\textbf{Comparison of Month} ~\figref{fig:ex_1} shows the forecast results by month.
The performance for light precipitation is similar regardless of month, but for the winter season (December and January), the CSI 1 mm performance decreases. 
This can be interpreted as a change in precipitation type with snowfall instead of rainfall in winter over the Korean peninsula. 
In addition, the reason the CSI 1 mm shows a relatively high in summer (June-August) is that there are more precipitation days than in winter, and precipitation events occur over a wide area during the monsoon season.
Heavy precipitation mainly occurs in summer, but it is not easy to predict because it occurs rarely, locally, and for a short time compared to light precipitation. 

\begin{figure*}[t!]
\centering
  \includegraphics[width=0.8\linewidth]{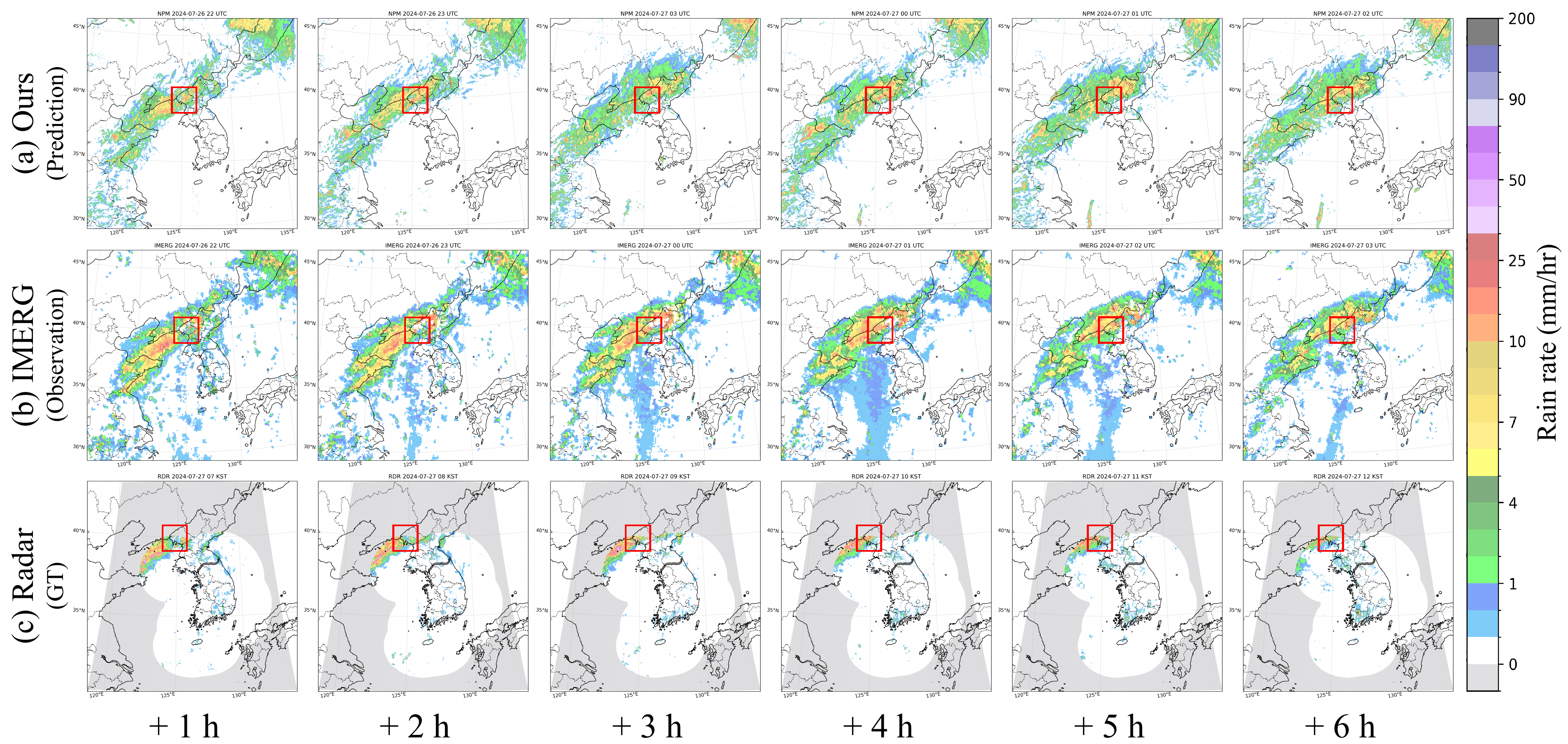}
  \caption{{Precipitation forecasting results of 2024-07-26 heavy rainfall case in North Korea. (a) is the prediction result of NPM, (b) is the global satellite-based precipitation data from NASA GPM IMERG-late run, and (c) is the observation data from KMA radar.}}
\label{fig:fig_north}
\end{figure*}
\noindent\textbf{Comparison of Image-to-Image Translation Models} ~\tableref{tab:I2I} presents the CSI scores according to the choice of image-to-image translation models used to predict satellite imagery.
We experiment with paired (aligned) GAN-based models such as Pix2Pix, BBDM, and unpaired (unaligned) models like StegoGAN, as well as StegoGAN trained in a paired setting (StegoGAN (Paired)).

The CSI scores indicate that StegoGAN (Paired) is the best choice in our framework, followed by the diffusion-based model, BBDM.
Although BBDM also achieves competitive scores, it requires 200 sampling steps, which means that StegoGAN (Paired) is more computationally efficient.
Note that our framework can operate with any Image-to-Image model. 

\noindent\textbf{Ablation Study} ~\tableref{tab:table_aa} presents the ablation study on each component in NPM.
We set the baseline of NPM as SimVP and conduct the experiment by progressively adding each component, including Large Kernel Attention (LKA), Temporal Large Kernel Attention (TKA), Temporal Consistency Constraint, Day Embedding, Hour Embedding, and Sampling Strategy.
As shown in~\tableref{tab:table_aa}, each component contributes to performance improvement, with Day Embedding being the most significant factor.
These results indicate that considering seasonal factors is crucial in the precipitation forecasting task.
Additionally, the MSE results of the video prediction models are as follows: PhyDNet (74.73), PredRNNv2 (99.25), SimVP (85.14), SimVP-V2 (84.22), TAU (70.20), and SwinLSTM (103.94), with NPM achieving 66.51.
These experimental results indicate that each component contributes to the improvement in the performance of the video prediction models.

\begin{figure}[t!]
\centering
  \includegraphics[width=\linewidth]{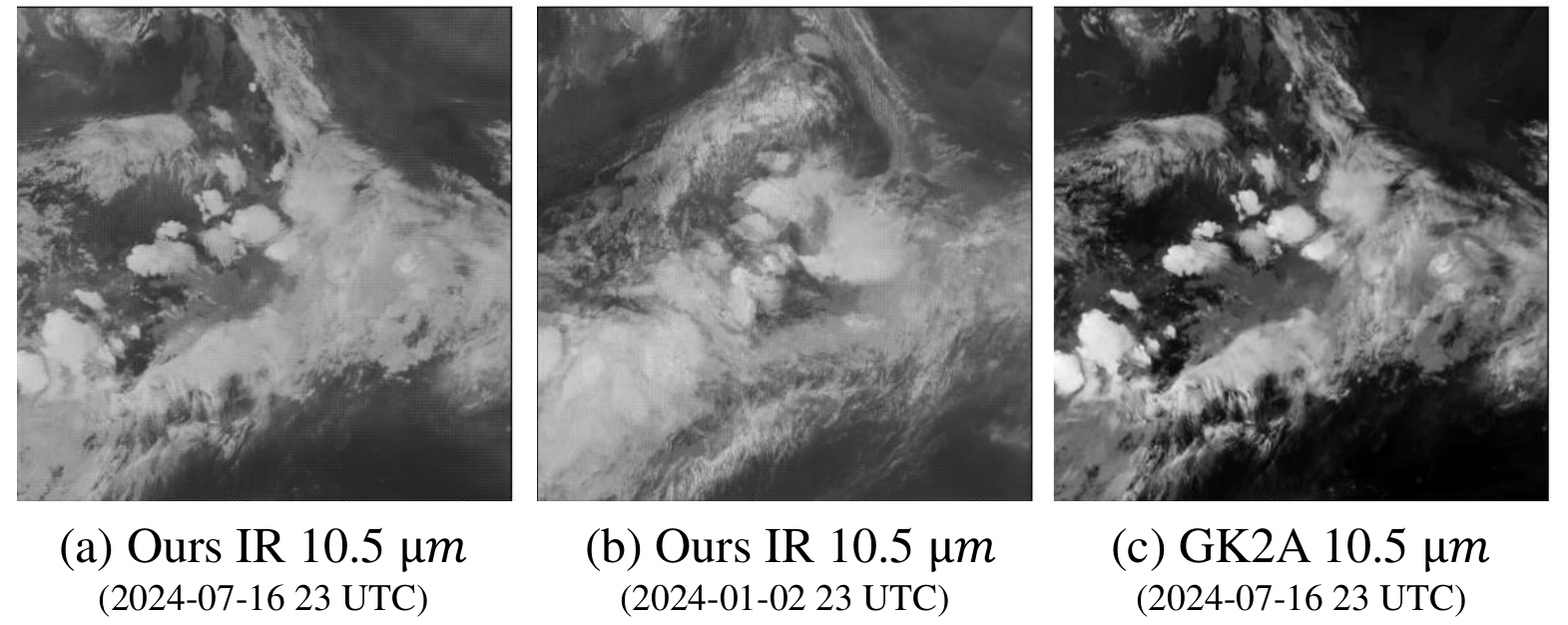}

  \caption{Comparison of Day embedding. (a) shows the results with the correct date input, (b) shows the results with only the date modified to 2024.01.02.2300 for the same input, and (c) is the ground truth (GT).}
\label{fig:fig5}
\end{figure}

\figref{fig:fig5} shows the result of our video prediction model and the role of Day embedding.
Our prediction result in~\figref{fig:fig5}-(a) show accurate predictions in terms of cloud morphology and location compared to the GT.
Also, when the Day embedding is changed from July to January with the same input, the prediction result shows different clouds, proving that the model reflects day conditions.
\subsection{Qualitative Results}

\figref{fig:fig_pray} shows a quantitative comparison between the Radar2Radar models, PreDiff and our Sat2Rdr model.
The cloud features (bright areas) in the IR 10.5 $\mu m$ channel indicate heavy precipitation signals.
However, the actual radar observations do not include a clear precipitation signal during the input time.
Consequently, as seen in~\figref{fig:fig_pray}-(b), the model that uses only radar input fails to generate future precipitation accurately when signals are not present in the current radar observations.

Additionally, PreDiff achieves an average performance of 0.14 for 1 mm precipitation over 6 hours. This is because PreDiff is designed to rely solely on radar.
In contrast, our NPM, which leverages cloud movement and generation context, is able to capture indirect signals related to precipitation onset and accurately predict the development of precipitation, even by considering factors that may not be directly observable.
Notably, the 6-hour average CSI of our approach is 0.29, demonstrating its ability to outperform PreDiff.
These experimental results indicate that Sat2Rdr has a significant advantage in predicting sudden-onset precipitation.

\subsection{Flood Case Study in North Korea }
To demonstrate the zero-shot capability of NPM for regions without radar observation, we conduct a case study on the July 2024 floods in North Korea as a part of our AI for Social Good initiative. The heavy rainfall near the Yalu River in North Korea caused levee breaches, leading to approximately 1,500 deaths or missing persons. Figure 5 illustrates the predictions generated by our model, along with South Korean radar observations and NASA GPM IMERG data analysis (noting that the IMERG-late run is accessible 14 hours later).

In~\figref{fig:fig_north}-(a), the red-colored box indicates the Yalu River in North Korea, where our model predicts significant rainfall over a continuous 6-hour period. ~\figref{fig:fig_north}-(b) shows that substantial rainfall is also recorded near the Yalu River by IMERG. However, IMERG data is available after a particular time, and that cannot be used for real-time disaster response.~\figref{fig:fig_north}-(c) shows the ground truths, where even more rainfall is observed compared to the NPM results. However, due to observation limitations, it is not possible to confirm the actual ground precipitation amount in the masked area.

In reality, the 6-hour accumulated rainfall in the region is approximately 60 mm, with IMERG recording 46.49 mm and NPM predicting 27.45 mm. Although the NPM tends to underestimate precipitation, it can capture accurate predictions of future 6-hour precipitation patterns and intensity.
These results suggest that NPM holds a potential for use in flood alerts in regions without any radar coverage.

\section{Conclusions}
In this paper, we address the limitations of existing data-driven precipitation forecasting methods, which either rely solely on radar data or are dependent on radar modalities.
To overcome this issue, we propose a precipitation prediction directly from satellite imagery, which facilitates auto-regressive forecasting by leveraging satellite-to-satellite video prediction and satellite-to-radar image-to-image translation methods. In addition, our dataset, named Sat2Radar, supports this task and allows us to fairly evaluate relevant methods including ours. 

Furthermore, we incorporate day and hour positional encoding to capture seasonal and time-dependent precipitation patterns.
We conduct a case study on a flood event in North Korea using our dataset and demonstrate the generality of our method in regions without anyy radar coverage.
We hope that our approach will be widely adopted in developing countries where radar installations are scarce.
In support of AI for social good, we will release our code and datasets in public.

\noindent\textbf{Limitation}
In this work, we aim to reduce the dependency on expensive hardware such as radar systems, but it still requires access to satellite imagery. 
This means that our method is not entirely free from reliance on high-cost equipment.
%
Nevertheless, it is important to note that satellites covering regions like GK2A (East Asia and Pacific), MSG4 (Europe, Africa), and GEOS18 (North and South America) provide free and easily accessible data.
\section{Acknowledgements}
This work was supported by the National Research Foundation of Korea (NRF) grant (RS-2024-00338439), and Institute of Information $\&$ communications Technology Planning $\&$ Evaluation (IITP) grant funded by the Korea government(MSIT) (RS-2021-II212068, Artificial Intelligence Innovation Hub and No.2019-0-01842, Artificial Intelligence Graduate School Program (GIST)). 

\bibliography{aaai25}

\newpage

\end{document}